\documentclass[runningheads]{llncs}
\usepackage[T1]{fontenc}

\usepackage{graphicx}
\usepackage{booktabs}
\usepackage{caption}
\usepackage{adjustbox}
\usepackage{xcolor}
\usepackage[nolist,nohyperlinks]{acronym}
\acrodef{DrCif}{Diverse Representation Canonical Interval Forest Classifier}
\acrodef{DTW}{Dynamic Time Warping}
\acrodef{EFDT}{Extremely Fast Decision Tree}
\acrodef{IDT}{Incremental Decision Tree}
\acrodef{IG}{Information Gain}
\acrodef{HAT}{Hoeffding Adaptive Tree}
\acrodef{HIVECOTE2}{Hierarchical Vote Collective of Transformation-based Ensembles 2.0}
\acrodef{HT}{Hoeffding Tree}
\acrodef{MIHT}{Multi-Instance Hoeffding Tree}
\acrodef{MIL}{Multi-Instance Learning}
\acrodef{MUSE}{Multivariate Symbolic Extension of WEASEL}
\acrodef{NN}{Nearest Neighbor}
\acrodef{ROCKET}{Random Convolutional Kernel Transform}
\acrodef{ST}{Shapelet Transform}
\acrodef{SVM}{Support Vector Machine}
\acrodef{TapNet}{Time series attentional prototype network}
\acrodef{TSC}{Time-Series Classification}
\acrodef{TPE}{Tree-structured Parzen Estimator}
\usepackage{tabularray}
\UseTblrLibrary{booktabs}
\usepackage{subfig}
\usepackage{comment}
\usepackage{amsmath,amsfonts,pifont}

\usepackage[top=1.64in,bottom=1in,left=1.88in,right=1.81in]{geometry}
\usepackage[vlined, ruled, linesnumbered]{algorithm2e}
\SetAlFnt{\small}
\DontPrintSemicolon
\SetKwProg{Fn}{Function}{:}{end function}
\SetKwFunction{FleafLearning}{leafLearning}

\makeatletter
\SetKwInput{KwSymbols}{Symbols}
\SetFuncSty{myitalic}
\SetCommentSty{myitalic}
\SetKwComment{Comment}{$\rhd~~$}
\makeatother

\begin{document}

\title{MIHT: A Hoeffding Tree for Time Series Classification using Multiple Instance Learning}

\titlerunning{MIHT: A Hoeffding Tree for TSC using MIL}

\author{Aurora Esteban
\and Amelia Zafra
\and Sebastián Ventura
}

\authorrunning{A. Esteban et al.}


\institute{Dept. of Computer Science and Artificial Intelligence, Andalusian Research Institute in Data Science and Computational Intelligence, University of Cordoba, Spain\\ \email{\{aestebant,azafra,sventura\}@uco.es}}

\maketitle              

\begin{abstract}
Due to the prevalence of temporal data and its inherent dependencies in many real-world problems, time series classification is of paramount importance in various domains. However, existing models often struggle with series of variable length or high dimensionality. This paper introduces the MIHT (Multi-instance Hoeffding Tree) algorithm, an efficient model that uses multi-instance learning to classify multivariate and variable-length time series while providing interpretable results. The algorithm uses a novel representation of time series as "bags of subseries," together with an optimization process based on incremental decision trees that distinguish relevant parts of the series from noise. This methodology extracts the underlying concept of series with multiple variables and variable lengths. The generated decision tree is a compact, white-box representation of the series' concept, providing interpretability insights into the most relevant variables and segments of the series. Experimental results demonstrate MIHT's superiority, as it outperforms 11 state-of-the-art time series classification models on 28 public datasets, including high-dimensional ones. MIHT offers enhanced accuracy and interpretability, making it a promising solution for handling complex, dynamic time series data.

\end{abstract}

\section{Introduction}

A time series, defined as a sequence of data points recorded over successive time intervals, proves to be a powerful tool to analyze trends, patterns, and dependencies in phenomena or variables over time \cite{Bagnall2017}. 
\ac{TSC} is a crucial machine learning task, emphasizing the extraction of temporal features to accurately assign labels to the entire series.
Time series data can be univariate, where a single variable is observed at each time point, or multivariate, with multiple variables recorded simultaneously at each timestamp. Moreover, time series within the same domain may have uniform or varying lengths. While several methods exist for \ac{TSC} in uniform-length contexts, especially for univariate time series \cite{Bagnall2017}, these approaches often involve step-by-step parsing, making them unsuitable for time series of unequal length. Additionally, they generate numerous attributes, leading to scalability challenges for long or multivariate time series. 
Recent findings \cite{Han2023,Wang2021} underscore the potential use of the \ac{MIL} framework for \ac{TSC}. \ac{MIL} \cite{Waqas2024}, a machine learning paradigm handling complex object problems by grouping instances into bags, offers a promising approach to managing the variability and uncertainty inherent in multivariate time series data of unequal length. 
Despite the limited attention received by \ac{MIL} in \ac{TSC}, existing approaches often combine it with complex generative models, such as auto-regression \cite{Ruiz-Munoz2015}, hidden Markov \cite{Guan2016}, or deep learning architectures \cite{Han2023}. However, these models face challenges like high computational complexity and a lack of interpretability, which limits their applicability for long-duration time-series analysis. Moreover, their application has been domain-specific, lacking a comprehensive exploration in diverse contexts compared to prevailing state-of-the-art \ac{TSC} methods. Hence, there is a pressing need to explore alternative MIL techniques that offer better scalability, interpretability, and applicability across a broader range of \ac{TSC} tasks.

In this paper, we introduce \acf{MIHT}, a novel \ac{MIL}-based approach for \ac{TSC}, addressing the above limitations through three key contributions. Firstly, we propose a multi-instance representation for time series, enabling effective learning from unequal-length series. This representation forms bags of instances from time-series subsequences, incorporating novel overlapping to enhance the capture of the concept characterizing the series label. Secondly, a novel optimization process identifies the most representative instances within the bag. The flexibility of this feature through different \ac{MIL} assumptions allows \ac{MIHT} to adapt to various domains with different concepts distributed across the series. Lastly, we pioneer the use of \acp{IDT} in \ac{TSC} as far as we know, by incorporating them as base learners of the \ac{MIL} framework. \acp{IDT} compound a family of algorithms designed to create and adapt the tree structure to changes in potentially infinite data streams. 
In combination with the proposed \ac{MIL} framework, they facilitate dynamic model updates, providing relevant predictions for multivariate time series of unequal length, while keeping low complexity and good performance in long time series. 

Section \ref{sec:rwork} gives a brief review of the state-of-the-art in multivariate \ac{TSC} in general and \ac{MIL}-based proposals in particular. Section \ref{sec:formulation} formally defines the characteristics of the problem. Section \ref{sec:proposal} presents our proposed \ac{MIHT}. Section \ref{sec:exp} tests the effectiveness of the proposal attending to both performance and interpretability. Section \ref{sec:conclusions} summarizes the conclusions obtained.

\section{Related Work}\label{sec:rwork}


\subsection{Multivariate Time Series Classification}\label{sec:rwork:tsc}

The development of \ac{TSC} models focuses mainly on uniform length series. Within this category, there are four fundamental approaches according to how they use the information of time series to perform the classification task \cite{Bagnall2017}: \textit{interval-based approaches} that divide the series into intervals and extract summary features from them; \textit{shapelet-based approaches} that find short patterns to characterize the series, \textit{dictionary-based approaches} that extract recurring patterns from the series, and \textit{random-convolutions-based methods}, that apply random convolutional kernels to extract features from the time series.
In each category, multiple proposals capable of dealing with multivariate series have been proposed. For example, DrCif \cite{Middlehurst2021} is one of the most recent interval-based approaches: it combines baseline summation with interval-based summation. The binary ST \cite{Bostrom2017} is the baseline for shapelet-based approaches for multivariate series. Attending to dictionary-based proposals, MUSE \cite{Schafer2017} is the most popular model, based on building bag-of-patterns using the Symbolic Fourier Approximation
. The exponent of methods based on random convolutions is ROCKET \cite{Dempster2020}, that uses convolutional kernels
.
Some proposals combine several of the mentioned categories in an ensemble, like HIVECOTE2 \cite{Middlehurst2021}, with very good results thanks to its heterogeneous optimization.
Multiple deep learning proposals have emerged as alternative to the traditional approaches. Two of the most popular models are TapNet \cite{Zhang2020} and InceptionTime \cite{Fawaz2020}, that combine CNNs, LSTMs, and attention mechanisms in a deep network for multivariate \ac{TSC}. 

Beyond uniform-length series, the main methods for variable-length \ac{TSC} are variations of \acp{SVM} and \ac{NN}. These proposals compare pairs of series by searching for the best alignment. In the case of \ac{SVM}, pairwise aggregators of typical kernels are used \cite{Faouzi2023}. For \ac{NN} different distances have been explored, like DTW \cite{Kate2016} that account for variations in the time axis to find the optimal alignment. 

\subsection{Multi-Instance Time Series Classification}

\ac{MIL} \cite{Waqas2024} is a learning paradigm characterized by grouping several instances in a single bag with the objective of finding the correct label for the whole bag. While very effective in representing complex objects, \ac{MIL} is not explicitly designed to deal with temporal data. There are few proposals for problem-specific \ac{TSC} in which different adaptations of \ac{MIL} to sequential data are proposed. These studies show that \ac{MIL} has great potential for real-world applications with time series with different lengths.
Thus, the simplest frameworks \cite{Stikic2011,Guan2016} form the bag considering each of the records as an instance, whereby the temporal information between records is ignored.
Other works \cite{Ruiz-Munoz2015,Janakiraman2018,Wang2021} preprocess the time series with convolutional filters or deep learning to extract features in a deeper dimensionality space and apply \ac{MIL} on them. It leads to an end-to-end classification model that provides little interpretability.
The most popular approach \cite{Dennis2018,Han2023,Chen2023} consists of splitting the series into regular sections and obtaining shorter subsequences.
In terms of the basic machine learning model, some proposals \cite{Stikic2011} use classical models adapted to \ac{MIL} such as Naïve Bayes or SVM, so that the temporal relationship is not explicitly learned. Other studies \cite{Wang2021,Han2023,Chen2023} use deep learning models based on CNNs, with the associated lack of interpretability. Finally, the proposals \cite{Guan2016,Janakiraman2018,Dennis2018} that work with subsequences of the series directly use autoregressive models, such as Hidden Markov models or LSTMs networks, which are also black-box models.


\section{Problem Formulation}\label{sec:formulation}



A multivariate time series $X = [\mathbf{x}_0,..., \mathbf{x}_l] \in \mathcal{R}^{l \times m}$ is an ordered sequence of $l \in \mathbb{N}$ steps where each step, $\mathbf{x}_i$, has exactly $m$ dimensions $[x_{i,0}, ..., x_{i,m}]$. This definition is extended when considering a set of $n$ time series $\mathcal{X} = \{X_0, ..., X_n\}$ extracted from the same process but with different durations. Thus, although all the series share $m$, each one has a different length $l$. 
In the \ac{TSC} problem, each time series has an associated output $y$ from a predefined set of classes $\mathcal{C} = \{c_1, ... c_n\}$. Thus, given a set of multivariate time series of variable length $\mathcal{X} \in \mathcal{R}^{n \times \max(l) \times m}$, our task is to learn the mapping function $f: \mathcal{X} \rightarrow \mathcal{C}$ that will allow us to predict $\hat{y}=f(X) \in \mathcal{C}$ for new $X \in \mathcal{X}$.

The time series for which we design this proposal have irregular length $l$ because the process that generates them has no fixed duration. Therefore, we start from the assumption that not all parts of the time series contribute to the final output of the process. We define the concept $\sigma$ of the series $X$ as the set of time steps $i \in l$ that encapsulate the information needed to predict $y$. 

\section{Proposed Model}\label{sec:proposal}

This section presents the complete specification of the proposed \acf{MIHT}
. Algorithm \ref{alg:miht} presents the general workflow: generating the multi-instance bags from the original time series (Section \ref{sec:prop:milseq}), initializing the \ac{IDT} (Section \ref{sec:prop:init}), iteratively searching for the $k$ most representative instances while reinforcing the model with them (Section \ref{sec:prop:bestk}), and using the trained model for class inference and interpretable insights (Section \ref{sec:prop:eval}).


\begin{algorithm}[!t]
\fontsize{7pt}{8pt}\selectfont 
    \caption{\acf{MIHT}}\label{alg:miht}
    
    \KwIn{\;
        $\mathcal{X} = \{(X_0, y_0), (X_1, y_1), ..., (X_n, y_n)\}$: set of training time series\;
        $\omega$: instance length\;
        $\lambda$: overlapping between instances\;
        $k$: number of consecutive instances to find\;
        \textit{Parameters of the base Hoeffding tree:}\;
        $\quad\delta$: significance level for the Hoeffding bound\;
        $\quad\kappa$: number of received instances a leaf should see before split attempts\;
    }
    \Comment{Create bags of instances:}
    \For{$i$ gets $0$ to $n$}{
        $l \leftarrow$ length of $X_i$\;
        $B_i \leftarrow ([\mathbf{i}_{i,j}, \mathbf{i}_{i,j+\omega}~($for $j$ gets $0$ to $l$ by $\lambda)],~y_i)$\;
    }
    \Comment{Initialize the base learner with all the instances of all the bags:}
    $HT \leftarrow$ new leaf\;
    \For{$B_i\in \mathcal{B}$}{
        \For{$\mathbf{i}_{i,j} \in B_i$}{
        $path \leftarrow$ sequence of nodes resulting from traverse $HT$ with $\mathbf{i}_{i,j}$\;
        \For{$node \in path$}{
            Update the statistics of $node$ based on $y_i$: $W$, $W_c$, $W_s$, $W_{c,s}$\;
            \If{$node$ is leaf and $\kappa$ instances received since last split attempt}{
                $\epsilon \leftarrow$ Hoeffding bound with tolerance $\delta$\; 
                Find the two splits $s_i$, $s_j$ that maximize $G_i$ and $G_j$\; 
                \If(\Comment*[f]{Meet the criterion for tree expansion}){$(G_i - G_j) > \epsilon$}{
                    \eIf{attribute $x$ associated to $s_i$ is categorical}{
                        Replace leaf by branch with conditions $x = s_i$, $x \neq s_i$\;
                    }{
                        Replace leaf by branch with conditions $x \leq s_i$, $x > s_i$\;
                    }
                    Traverse $X_i$ until new leaf $\in$ $path$ and update stats\;
                }
            }
        }
        }
    }
    \Comment{Optimize to the best k consecutive instances of each bag:}
    \While{not convergence}{
        \For{$B_i\in \mathcal{B}$}{
            For each $\mathbf{i}_{i,j} \in B_i$, traverse tree and get $\mathcal{L}$ in resulting leaf (Eq. \ref{eq:likelihood})\;
            $\tau_i \leftarrow \arg \max \sum_{j' \geq j > j'+k} \mathcal{L}(\mathbf{i}_{i,j'}, y_i)~\forall j \in B_i$ (Eq. \ref{eq:optimizationfunc})\;
            Repeat procedure of lines 7-23 only with $\tau_i$ instead of the whole $B_i$\;
        }
    }
\end{algorithm}

\subsection{Representing time series as multi-instance bags}\label{sec:prop:milseq}

\ac{MIL} \cite{Waqas2024} is a framework for representing complex objects by grouping instances belonging to the same object under a bag $B$. 
In traditional \ac{MIL}, each bag is a set of a variable number of point-based instances that do not keep temporal information. To address the \ac{TSC} problem, we propose a novel \ac{MIL} representation
. Here, the original time series $X$ is transformed into a bag $B = [\mathbf{i}_0,..., \mathbf{i}_n]$ as an ordered succession of subseries. Each subseries $\mathbf{i}_i = [\mathbf{x}_j, ... \mathbf{x}_{j+\omega}]$ is obtained by applying a sliding window to $X$ of size $\omega$ and overlap $\lambda$, taking time steps such that  $[x_i, ... x_{i+\omega}]~ \forall i \in set(1, 1+\lambda, 1+2\lambda, 1+3\lambda, … l )$. In this way, we pass from the previously defined set of sequences $\mathcal{X}$ to a set of bags $\mathcal{B} = \{B_0, ..., B_n\} \in \mathcal{R}^{n \times \omega \times \max(n) \times m}$, where each bag is composed by a variable number of instances $n$ of $\omega$ steps, with each step of $m$ dimensions.
On the other hand, the output space of the original \ac{TSC} problem $\mathcal{C}$ has not changed (see Section \ref{sec:formulation}). Thus, the objective of the multi-instance \ac{TSC} task is still to build the function $f: \mathcal{B} \rightarrow \mathcal{C}$ to predict a class label for each bag, i.e., for each time series. This transformation of the input space occurs in lines 1-3 of Algorithm \ref{alg:miht}.


As part of this transformation, the inner concept $\sigma$ of the series will be spread across certain instances. We define $\tau \in B$ as the set of $k$ instances that better approximate $\sigma$ for a given bag $B$. This allows us to propose an optimization process for $\tau$ as the instances that maximize the bag class likelihood $\mathcal{L}$:

\begin{equation}\label{eq:optimizationfunc}
    \text{argmax}_{\mathbf{i}_j, j \in [0,n)} \sum_{j,\tau} \gamma_{j,\tau} \mathcal{L}(\mathbf{i}_{j}, y) ~ | \gamma_{j,\tau} =
    \begin{cases}
        1 ~\forall \tau \in [\mathbf{i}_{j}, \mathbf{i}_{j+k}) \\
        0, ~\text{ otherwise} \\
    \end{cases}
\end{equation}

The computation of $\mathcal{L}$ depends on the base learner that processes the instances in each bag. In this line, this model must be incremental, i.e., it must be able to update its internal model throughout the optimization process of $\tau$ without forgetting past information when it comes to an iterative search. This is why we choose \acp{IDT} as we explain in the next section.

In summary, our proposal assumes that not the whole time series is useful for its assigned class in the context of \ac{TSC}, but that the information is in $\sigma$. The proposed \ac{MIL}-based representation gives us the framework to isolate $\sigma$ by an optimization process that discriminates the $k$ instances containing it, $\tau$. $\sigma$ depends on the problem and often also on each time series; therefore, $\omega$, $\lambda$ and $k$ are parameters that make \ac{MIHT} adaptable to multiple domains. 


\subsection{Initializing the base learner}\label{sec:prop:init}

\acp{IDT} are characterized by dynamically building the tree model as information becomes available, being able to adapt its structure to new data without forgetting previous knowledge and reducing the need for exhaustive exploration. The \ac{HT} \cite{Domingos2000} is a remarkable \ac{IDT} that relies on the Hoeffding bound to make early decisions on attribute splits with statistically significant confidence. The Hoeffding bound $\epsilon$ provides statistical guarantees about the relationship between the true mean and observed mean of independent random variables –in our problem, the instances in $\mathcal{B}$. 
This statistical foundation enables a split criterion: when evaluating potential splits at a leaf node, the tree compares the difference in quality measures (typically \acf{IG}) between the two best splitting attributes. If this difference exceeds $\epsilon$, \ac{HT} concludes with confidence $1 - \delta$ that the best attribute is indeed statistically superior
.

With this foundation, \ac{MIHT} initializes the base \ac{HT} as detailed in Algorithm \ref{alg:miht} (lines 5-18). Starting with a single leaf node, the model processes all the instances of each bag as $m \times \omega$ vectors, where $m$ represents the time series dimensions and $\omega$ the instance length. The tree grows by evaluating each attribute $x$ every $\kappa$ instances, creating splits when the Hoeffding bound criterion is met, with leaf nodes updating their class distributions as new instances arrive.

A bag may contain instances that do not belong to the true concept $\sigma$. To address this, the $HT$ creates splits when sufficient instances demonstrate consistent distribution patterns. 
During initial learning using $(\mathbf{i}_{i,j}, y_i)$ tuples, instances belonging to $\tau$ naturally achieve higher \ac{IG}. While the distinction between relevant instances and noise is initially subtle due to limited data, the iterative process enables the tree to refine its predictions as more instances arrive at nodes, surpassing the Hoeffding bound. This refinement process specifically requires an \ac{IDT}, as traditional trees cannot calculate \ac{IG} online, lacking the ability to build and refine the tree structure incrementally as instances are processed.


\subsection{Optimizing the relevant instances}\label{sec:prop:bestk}

After initial \ac{HT} training on the complete dataset, a reinforcement phase begins (Algorithm \ref{alg:miht}, lines 19-23) to optimize both the \ac{HT} and the set of relevant instances $\tau_i$ for each bag $B_i \in \mathcal{B}$. This iterative process involves two key steps: i) selecting $k$ instances that maximize the likelihood of the bag's class $c \in \mathcal{C}$, and then ii) retraining the \ac{HT} with this refined dataset. This process, similar to the initial training in Section \ref{sec:prop:init} but with a more evolved tree, progressively approximates the true series concept $\sigma$ through increasingly focused data distributions.

For step (i), each instance's likelihood $\mathcal{L}$ of belonging to its bag class $y_i$ is determined through tree traversal enhanced by Adaptive Naive Bayes for probabilistic inference. As an instance $i$ traverses the tree to reach a leaf node, the node's accumulated statistics 
are used in a Naive Bayes calculation to determine the probability of the instance being classified as its actual bag class:

\begin{equation}\label{eq:likelihood}
    \mathcal{L} = p(y_i = c | \mathbf{i}) = \frac{p(\mathbf{i} | y_i = c) p(y_i = c)}{\sum_{c_i \in \mathcal{C}} p(\mathbf{i} | c_i) p(c_i)}
\end{equation}
where the class priors $p(c)$ are defined based on the prevalence of $c$ in the seen instances in the node, and the conditional probability of the instance given the class $c$ are obtained per each of its $m$ attributes such that $p(\mathbf{i} | c) = \sum_{j=1}^m p(x_j | c)$. 


For step (ii), the optimization process selects $k$ instances per bag that maximize the bag class likelihood using Equation (\ref{eq:optimizationfunc}), which are then used for incremental \ac{HT} training. 
MIHT's design enables model adaptation to this iterative process despite uncertainty in iteration count and data distribution changes. Each iteration refines the model toward $\tau$ selection while maintaining interpretability through a single tree that preserves only the relevant branches.

\subsection{Providing classifications}\label{sec:prop:eval}

The prediction process, detailed in Algorithm \ref{alg:prediction}, 
begins by transforming the unknown series into a multi-instance bag using the same procedure applied during training (described in Section \ref{sec:prop:milseq}). Each instance $i$ in the bag then traverses the \ac{HT} until reaching a leaf node. At this point, the instance's classification $\hat{y}$ is determined by maximizing the likelihood function (Equation \ref{eq:likelihood}) at the leaf node
.

The second step is to pass from the list of predictions at the instance level $[\hat{y'}_0, ... \hat{y'}_n]$ to the bag prediction $\hat{y}$. 
In order to consider the multi-class scenario, we employ the statistic mode, that states that the bag class is the most repeated label among the bag instances.
Finally, the set of instances most closely related to the predicted class $\hat{\tau}$ is estimated, which would be equivalent to the series concept $\sigma$ that gives information on when the event that characterizes the time series occurred. To do so, we again resort to the likelihood of belonging of all instances in the bag to the predicted class $\hat{y}$, and select the $k$ instances that maximize it
.

\begin{algorithm}[t]
\fontsize{7pt}{8pt}\selectfont 
    \caption{Classification with \ac{MIHT}}\label{alg:prediction}

    \KwIn{\;
        $X$: an unlabeled time series of undetermined length and same domain than $\mathcal{X}$\;
    }

    \Comment{Create the bag of instances:}
    $t \leftarrow$ length of $X$\;
    $B \leftarrow [\mathbf{i}_{j}, \mathbf{i}_{j+\omega}~($for $j$ gets $0$ to $t$ by $\lambda)]$\;
    
    \Comment{Getting the predicted class for $X$:}
    $\mathbf{\hat{y'}} \leftarrow [\hat{y'}_{\mathbf{i}} ~ \forall \mathbf{i} \in B],~~\hat{y'} = \text{argmax}_{c \in \mathcal{C}} ~ p(y' = c | \textbf{i}) $ 
    \;
    $\hat{y} \leftarrow$ MIL-assumption$(\mathbf{\hat{y'}})$\;

    \Comment{Most relevant instances for $\hat{y}$:}
    $\hat{\tau} \leftarrow \arg \max \sum_{i' \geq i > i'+k} \mathcal{L}(\mathbf{i}_{i'}, \hat{y})~\forall i \in B$ (Eq. \ref{eq:optimizationfunc})\;
\end{algorithm}

\section{Experimental Study}\label{sec:exp}

Section \ref{sec:exp:setup} establishes the framework followed for all the experiments. 
Section \ref{sec:exp:perform} studies the accuracy of the proposal against the state-of-the-art in \ac{TSC}. Finally, Section \ref{sec:exp:interpretability} analyzes \ac{MIHT}'s interpretability. 

\subsection{Experimental Setup}\label{sec:exp:setup}


MIHT focuses on variable-length, multivariate time series, where there is not as much work as in the case of uniform-length \ac{TSC}. Thus, the proposed method is evaluated on 28 datasets that compose the set of time series of variable length and multiple dimensions available in the public UCR/UEA archive for \ac{TSC} \cite{Bagnall2017}. These datasets balance the proportion between multivariate and univariate time series in the experimentation. The complete specification is available in Table \ref{tab:datasets}. 

\begin{table}[t]
    \centering
    \caption{Datasets description}\label{tab:datasets}
    \begin{tblr}{
    	width=\linewidth,
        colspec={*{2}{X[l,m] *{4}{Q[r,m]} Q[c,m]}},
        rowsep=-1pt,
        colsep=2pt,
        row{1-Z}={font=\fontsize{6}{6}},
    }
\toprule
Dataset & Vars & Class & {Train\\size} & {Test\\size} & {Series\\length} & Dataset & Vars & Class & {Train\\size} & {Test\\size} & {Series\\length}\\
\midrule
AsphaltObstac & 1 & 4 & 390 & 391 & 298.6$\pm$114.5 & AsphObstCoor & 3 & 4 & 390 & 391 & 298.6$\pm$114.5 \\
AsphPavType & 1 & 3 & 1055 & 1056 & 400.2$\pm$168.2 & AsphPavTCoor & 3 & 3 & 1055 & 1056 & 400.2$\pm$168.2 \\
AsphRegular & 1 & 2 & 751 & 751 & 384.0$\pm$230.2 & AsphRegulCoor & 3 & 2 & 751 & 751 & 384.0$\pm$230.2\\
AllGestWiimX & 1 & 10 & 300 & 700 & 124.7$\pm$68.0 & CharactTraject & 3 & 20 & 1422 & 1436 & 119.9$\pm$21.1\\
AllGestWiimY & 1 & 10 & 300 & 700 & 124.7$\pm$68.0 & InsectWingbeat & 200 & 10 & 25k & 25k & 6.7$\pm$1.6\\
AllGestWiimZ & 1 & 10 & 300 & 700 & 124.7$\pm$68.0 & JapaneseVowels & 12 & 9 & 270 & 370 & 15.6$\pm$3.6 \\
GestMidAirD1 & 1 & 26 & 208 & 130 & 167.5$\pm$63.2 & ArticulWordRec & 9 & 25 & 275 & 300 & 144.0$\pm$0.0\\
GestMidAirD2 & 1 & 26 & 208 & 130 & 167.5$\pm$63.2 &  SpokenArabDig & 13 & 10 & 6599 & 2199 & 39.8$\pm$8.6\\
GestMidAirD3 & 1 & 26 & 208 & 130 & 167.5$\pm$63.2 & AtrialFibr & 2 & 3 & 15 & 15 & 640.0$\pm$0.0\\
GestPebbleZ1 & 1 & 6 & 132 & 172 & 219.3$\pm$75.3 & ERing & 4 & 6 & 30 & 270 & 65.0$\pm$0.0\\
GestPebbleZ2 & 1 & 6 & 146 & 158 & 219.3$\pm$75.3 & HandMovemDir & 10 & 4 & 160 & 74 & 400.0$\pm$0.0\\
PickupGestWiim & 1 & 10 & 50 & 50 & 145.7$\pm$73.7 & Heartbeat & 61 & 2 & 204 & 205 & 405.0$\pm$0.0\\
PLAID & 1 & 11 & 537 & 537 & 325.4$\pm$140.8 & SelfRegulSCP2 & 7 & 2 & 200 & 180 & 1152.0$\pm$0.0\\
ShakeGestWiimZ & 1 & 10 & 50 & 50 & 167.3$\pm$86.9 & StandWalkJump & 4 & 3 & 12 & 15 & 2500.0$\pm$0.0\\
\bottomrule
\end{tblr}
\end{table}


Classic metrics of the classification paradigm were considered to evaluate the performance of our proposed \ac{MIHT}: the accuracy, balanced accuracy, and hamming loss to measure the performance at instance level, and the F1-score micro and macro-averaged for measuring of the performance at class-level.


The comparative study methods cover the main categories of the current state-of-the-art for multivariate \ac{TSC} (see Section \ref{sec:rwork:tsc}). For methods that only support fixed-length \ac{TSC}, the variable-length problems have been truncated to their shortest series.
The parameters for each model are shown in Table \ref{tab:sota} and  correspond to the values recommended by the authors, with no individual adjustment per problem in any case. In the case of \ac{MIHT}, the parameters have been established based on a heuristic search using a subset of the training partitions of the datasets included in the experimentation
. The source code of the proposal is publicly available in the repository associated\footnote{Additional materials available at \url{https://github.com/aestebant/miht}}, together with the instructions for reproducible experimentation. 

\begin{table}[t]
    \centering
    \caption{TSC comparison methods}\label{tab:sota}
    \begin{tblr}{
    	width=\linewidth,
        colspec={Q[l,m] Q[c,m] X[l,m]},
        rowsep=-1pt,
        colsep=2pt,
        row{1-Z}={font=\fontsize{6}{6}},
        cell{2-Z}{3}={font=\itshape\fontsize{6}{6}},
    }
        \toprule
        Method & Family & Parameters\\
        \midrule
        DrCif \cite{Middlehurst2021} & {Feature\\based} & intervals: $4+(\sqrt{length} \times \sqrt{vars})/3$, base estimator: CIT, estimators 200\\
        ST \cite{Bostrom2017} & Shapelet & shapelet candidates: 10000, base estimator: DT, estimators 200\\
        MUSE \cite{Schafer2017} & Dict. & coeff select: ANOVA, bigrams: True, window inc: 2, alphabet size: 4, feature select: $\chi^2$\\
        {ROCKET\cite{Dempster2020}} & {Convol.} & kernels: 10000, max dilatation per kernel: 32, features per kernel: 4\\
        {SVM-\\Linear \cite{Loning2019}} & {Kernel\\based} & kernel: linear, regularization: 1.0, shrinking: True, stopping tolerance: 1e-3, decision func: ovr, max iterations: 100\\
        {SVM-\\RBF \cite{Loning2019}} & {Kernel\\based} & kernel: RBF, regularization: 1.0, shrinking: True, stopping tolerance: 1e-3, decision func: ovr, max iterations: 100\\
        {kNN-ED \cite{Loning2019}} & {Distance} & func: Euclidean distance, k: 1, weights: uniform, search: brute\\
        {kNN-DTW \cite{Kate2016}} & {Distance} & func: \acl{DTW}, k: 1, weights: uniform, search: brute\\
        TapNet \cite{Zhang2020} & {Deep\\learning} & components: rand projections + attention + LSTM + CNN, filter sizes: (256, 256, 128), kernel size: (8, 5, 3), layers: (500, 300), dropout: 0.5, epochs: 100\\
        {Inception\\Time \cite{Fawaz2020}} & {Deep\\learning} & kernel: 40, filters: 32, use residuals, bottleneck size: 32, depth: 6, epochs: 100\\
        {HIVE\\COTE2 \cite{Middlehurst2021}} & Hybrid & ST params: shapelet candidates: 10000, base estimator: DT, estimators 200. DrCif params: intervals: $4+(\sqrt{length} \times \sqrt{vars})/3$, base estimator: CIT, estimators 200. ARSENAL params: transform: ROCKET, estimators: 25. TDE params: combinations: 250, max ensemble size: 50, min win len: 10, selected params: 50\\ 
        MIHT & MIL+IDT & inst length $\omega$: 21\% of the series, inst overlap $\lambda$: 2\%, k: 4, $\kappa$: 366.5\%, $\delta$: 0.005615, split criterion: IG, max iterations: 100\\
        \bottomrule
    \end{tblr}
\end{table}






\subsection{Classification Performance}\label{sec:exp:perform}

This section evaluates the performance of \ac{MIHT} against the state-of-the-art \ac{TSC} on variable length time series
. The study starts with a detailed analysis attending to the accuracy metric, that is representative enough given the general balance between classes in UCR/UEA repository \cite{Middlehurst2021,Dempster2020,Fawaz2020}, to identify differences according to the characteristics of the datasets in the performance of the algorithms. Then, we carry out a more global analysis, analyzing average test results for all the metrics and using statistical tests to identify significant differences between algorithms. The complete results are available in the repository associated.

\begin{table}[!t]
    \centering
    \caption{Accuracy results on test data for MIHT and state-of-the-art methods}\label{tab:detailres}
        
    \begin{tblr}{
        width=\linewidth,
        colspec={Q[m,l] *{12}{X[m,r]}},
        rowsep=-1pt,
        colsep=2pt,
        row{1-Z}={font=\fontsize{6}{6}},
    }
        \toprule
        & MIHT & DrCIF & ST & MUSE & {ROC\\KET} & {SVM\\Linear} & {SVM\\RBF} & {kNN\\ED} & {kNN\\DTW} & {Tap\\Net} & {Incept\\time} & {HIVEC\\OTE2} \\
        \midrule
AsphObst & 0.611 & 0.609 & 0.545 & 0.555 & 0.540 & 0.266 & 0.355 & 0.297 & 0.488 & 0.522 & 0.545 & \textbf{0.619} \\
AsphPavT & 0.791 & 0.785 & 0.731 & 0.688 & 0.679 & 0.374 & 0.745 & 0.428 & 0.617 & 0.788 & \textbf{0.810} & 0.781 \\
AsphReg & 0.940 & 0.917 & 0.871 & 0.732 & 0.748 & 0.514 & 0.916 & 0.487 & 0.750 & 0.549 & \textbf{0.941} & 0.915 \\
AllGeWX & \textbf{0.363} & 0.213 & 0.137 & 0.126 & 0.137 & 0.130 & 0.273 & 0.193 & 0.197 & 0.174 & 0.171 & 0.209 \\
AllGeWY & \textbf{0.331} & 0.254 & 0.140 & 0.150 & 0.236 & 0.171 & 0.243 & 0.253 & 0.250 & 0.250 & 0.190 & 0.254 \\
AllGeWZ & 0.294 & \textbf{0.341} & 0.249 & 0.240 & 0.259 & 0.167 & 0.237 & 0.279 & 0.316 & 0.236 & 0.279 & 0.320 \\
GMAirD1 & 0.154 & 0.369 & 0.369 & 0.408 & 0.408 & 0.092 & 0.131 & 0.254 & 0.223 & 0.315 & 0.285 & \textbf{0.431} \\
GMAirD2 & 0.115 & \textbf{0.346} & 0.262 & 0.285 & 0.285 & 0.054 & 0.100 & 0.200 & 0.215 & 0.231 & 0.254 & 0.315 \\
GMAirD3 & 0.038 & 0.185 & 0.223 & 0.192 & 0.208 & 0.015 & 0.115 & 0.123 & 0.108 & 0.138 & 0.185 & \textbf{0.238} \\
GPebbleZ1 & \textbf{0.698} & 0.529 & 0.459 & 0.535 & 0.448 & 0.448 & 0.599 & 0.314 & 0.320 & 0.256 & 0.244 & 0.483 \\
GPebbleZ2 & \textbf{0.639} & 0.316 & 0.373 & 0.367 & 0.348 & 0.418 & 0.310 & 0.259 & 0.272 & 0.316 & 0.228 & 0.361 \\
PickGWZ & \textbf{0.540} & \textbf{0.540} & 0.460 & 0.400 & 0.320 & 0.360 & 0.420 & 0.460 & 0.400 & 0.240 & 0.100 & 0.460 \\
PLAID & 0.322 & 0.778 & 0.786 & 0.851 & 0.827 & 0.263 & 0.291 & 0.547 & \textbf{0.855} & 0.298 & 0.479 & 0.842 \\
ShakGWZ & 0.620 & 0.580 & 0.580 & 0.560 & 0.520 & 0.440 & \textbf{0.680} & 0.620 & 0.500 & 0.220 & 0.320 & 0.520 \\
AsphOC & \textbf{0.624} & 0.573 & 0.570 & 0.509 & 0.601 & 0.284 & 0.427 & 0.258 & 0.417 & - & - & 0.601 \\
AsphPTC & 0.819 & 0.871 & 0.816 & 0.806 & 0.827 & 0.381 & 0.782 & 0.449 & 0.536 & - & \textbf{0.885} & 0.868 \\
AsphRegC & \textbf{0.948} & 0.937 & 0.892 & 0.885 & 0.876 & 0.501 & 0.912 & 0.248 & 0.582 & 0.678 & - & 0.947 \\
CharTraj & 0.908 & 0.895 & 0.854 & 0.880 & 0.895 & 0.589 & 0.900 & 0.700 & \textbf{0.912} & 0.811 & - & 0.900 \\
InseWbeat & \textbf{0.381} & - & - & 0.222 & 0.187 & 0.126 & 0.113 & 0.165 & 0.254 & - & - & - \\
JapanVow & 0.911 & 0.900 & 0.568 & 0.435 & 0.873 & 0.957 & \textbf{0.968} & 0.343 & 0.900 & - & - & 0.900 \\
SpArabD & \textbf{0.933} & - & 0.384 & 0.608 & 0.636 & 0.524 & 0.483 & 0.486 & 0.652 & - & 0.691 & - \\
ArtWordR & 0.980 & 0.980 & 0.967 & \textbf{0.993} & \textbf{0.993} & 0.053 & 0.813 & 0.970 & 0.987 & 0.903 & 0.987 & \textbf{0.993} \\
AtrialFibr & \textbf{0.400} & 0.200 & 0.267 & 0.200 & 0.067 & 0.333 & 0.267 & 0.267 & 0.200 & 0.200 & 0.333 & 0.200 \\
ERing & 0.922 & \textbf{0.989} & 0.952 & 0.959 & 0.981 & 0.207 & 0.833 & 0.944 & 0.915 & 0.574 & 0.889 & \textbf{0.989} \\
HandMD & 0.459 & 0.486 & 0.392 & 0.243 & \textbf{0.527} & 0.311 & 0.189 & 0.257 & 0.189 & 0.243 & 0.405 & 0.473 \\
Heartbeat & 0.741 & \textbf{0.771} & 0.746 & 0.741 & 0.746 & 0.288 & 0.722 & 0.620 & 0.717 & 0.585 & 0.639 & 0.727 \\
SelfRSCP2 & 0.556 & 0.550 & 0.506 & \textbf{0.583} & 0.528 & 0.528 & 0.500 & 0.483 & 0.528 & 0.494 & 0.506 & 0.578 \\
StandWJ & \textbf{0.667} & 0.400 & 0.533 & 0.400 & 0.467 & 0.467 & 0.200 & 0.200 & 0.200 & 0.267 & 0.333 & 0.333 \\
        \bottomrule
   \SetCell[c=13]{l,\linewidth} Best results are in bold. 
   '-' indicates that the experiment failed due to the characteristics of the dataset or did not finish in the maximum time per experiment (480 hours).\\
    \end{tblr}
\end{table}

Table \ref{tab:detailres} presents accuracy results by model and dataset. While no single algorithm excels across all datasets due to their diverse characteristics, MIHT demonstrates robust performance, achieving best results in 11 out of 28 datasets, particularly excelling in high-dimensional data (60-200 attributes) where complex algorithms face scalability challenges.
Other models show dataset-specific strengths: MUSE performs well on uniform-length series with fewer variables, while deep learning approaches like InceptionTime and TapNet require substantial data and struggle with larger multivariate series. Traditional approaches using features, shapelets, or dictionaries show reduced effectiveness with variable-length series, while KNN-DTW's alignment process is less effective beyond uniform-length problems. Complex learning models (DrCif, HIVECOTE2) fail to complete processing on the largest datasets. 
MIHT's versatility stems from its k-mechanism, which regulates the analysis of consecutive instances, and its MIL paradigm, which offers flexible hypotheses for instance-label relationships. This adaptability enables effective concept discovery across diverse problems
.

Table \ref{tab:averes} summarizes the performance metrics across all methods. MIHT achieves superior results across all metrics, showing approximately 17\% improvement over non-ensemble methods like SVM and NN-based approaches. Ensemble-based methods such as DrCif, ST, and HIVECOTEV2 perform closer to MIHT, with approximately 5\% lower average performance.
Statistical analyses using the nonparametric Friedman test show that MIHT consistently achieves the highest rankings across all metrics, with p-values below 0.01 indicating significant differences at 99\% confidence level. The subsequent Bonferruni-Dunn post-hoc test results in Figure \ref{fig:posthoc} further delineate the performance groupings among the different models.
This analysis reveals that MIHT significantly outperforms single-model approaches of comparable complexity, such as kNN-DTW and SVM-RBF, as well as their simpler variants (kNN-ED, SVM-Linear) which are commonly used for uniform-length time series classification. Deep learning models, particularly TapNet, demonstrate significantly lower performance than MIHT, suggesting limited effectiveness in temporal pattern extraction in small datasets, or facing scalability challenges with high-dimensional series.
ROCKET and MUSE, using convolution-based and dictionary-based approaches respectively, show strong performance while maintaining computational simplicity. Their effectiveness with variable-length series stems from their ability to identify critical information at different temporal locations rather than fitting the entire series.
While ensemble-based models achieve comparable performance to MIHT through their combination of hundreds of decision trees, the differences are not statistically significant. MIHT maintains a slight average performance advantage with superior interpretability characteristics, as discussed in Section \ref{sec:exp:interpretability}. 

\begin{adjustbox}{minipage=0.48\textwidth,valign=b}
\captionof{table}{Average performance on test data, and Friedman results per metric}\label{tab:averes}
\begin{tblr}{
width=\linewidth,
colspec={Q[l,m] *{5}{X[r,m]}},
rowsep=-3pt,
colsep=1pt,
row{1-Z}={font=\tiny},
}
\toprule
 & Acc & Balanced acc & Hamming loss & Macro F1 & Micro F1\\
\midrule
MIHT & \textbf{0.5967} & \textbf{0.5820} & \textbf{0.4033} & \textbf{0.5706} & \textbf{0.5967} \\ 
DrCIF & 0.5470 & 0.5358 & 0.4530 & 0.5298 & 0.5470 \\ 
ST & 0.5225 & 0.5092 & 0.4775 & 0.5009 & 0.5225 \\ 
MUSE & 0.5198 & 0.5069 & 0.4802 & 0.4983 & 0.5198 \\ 
ROCKET & 0.5416 & 0.5303 & 0.4584 & 0.5215 & 0.5416 \\ 
SVM-Lin & 0.3307 & 0.3304 & 0.6693 & 0.2763 & 0.3307 \\ 
SVM-RBF & 0.4831 & 0.4659 & 0.5169 & 0.4204 & 0.4831 \\ 
kNN-ED & 0.3966 & 0.3889 & 0.6034 & 0.3668 & 0.3966 \\ 
kNN-DTW & 0.4821 & 0.4732 & 0.5179 & 0.4649 & 0.4821 \\ 
TapNet & 0.3317 & 0.3315 & 0.6683 & 0.2970 & 0.3317 \\ 
IncTime & 0.3821 & 0.3800 & 0.6179 & 0.3525 & 0.3821 \\ 
HCV2 & 0.5449 & 0.5296 & 0.4551 & 0.5212 & 0.5449 \\
\midrule
$\chi^2$ & 86.772 & 79.733 & 86.772 & 93.867 & 86.772\\
$p$-value & 7.13e-14 & 1.66e-12 & 7.13e-14 & 2.90e-15 & 7.13e-14\\
\bottomrule
\SetCell[c=6]{l,\linewidth} Best results are in bold.\\
\end{tblr}

\end{adjustbox}
\hfill
\begin{adjustbox}{minipage=0.48\textwidth,valign=b}
    \centering
    \includegraphics[width=\linewidth]{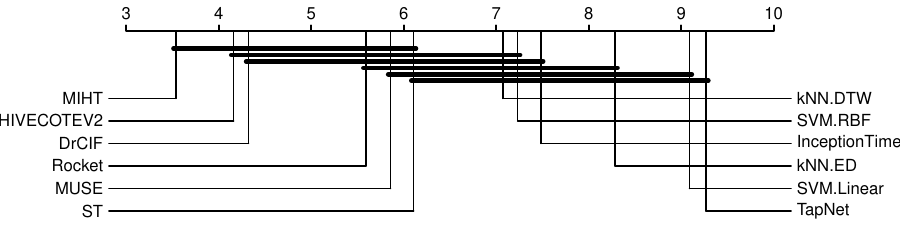}
    \captionof{figure}{Significant differences according to Bonferroni-Dunn's post-hoc test for Accuracy, $\alpha = 99\%$, over Friedman's rankings}
    \label{fig:posthoc}
\end{adjustbox}


\subsection{Interpretability Analysis}\label{sec:exp:interpretability}

\begin{figure}[t]
	\centering
    \subfloat[Decision tree generated \label{fig:tree}]{\includegraphics[width=0.4\linewidth]{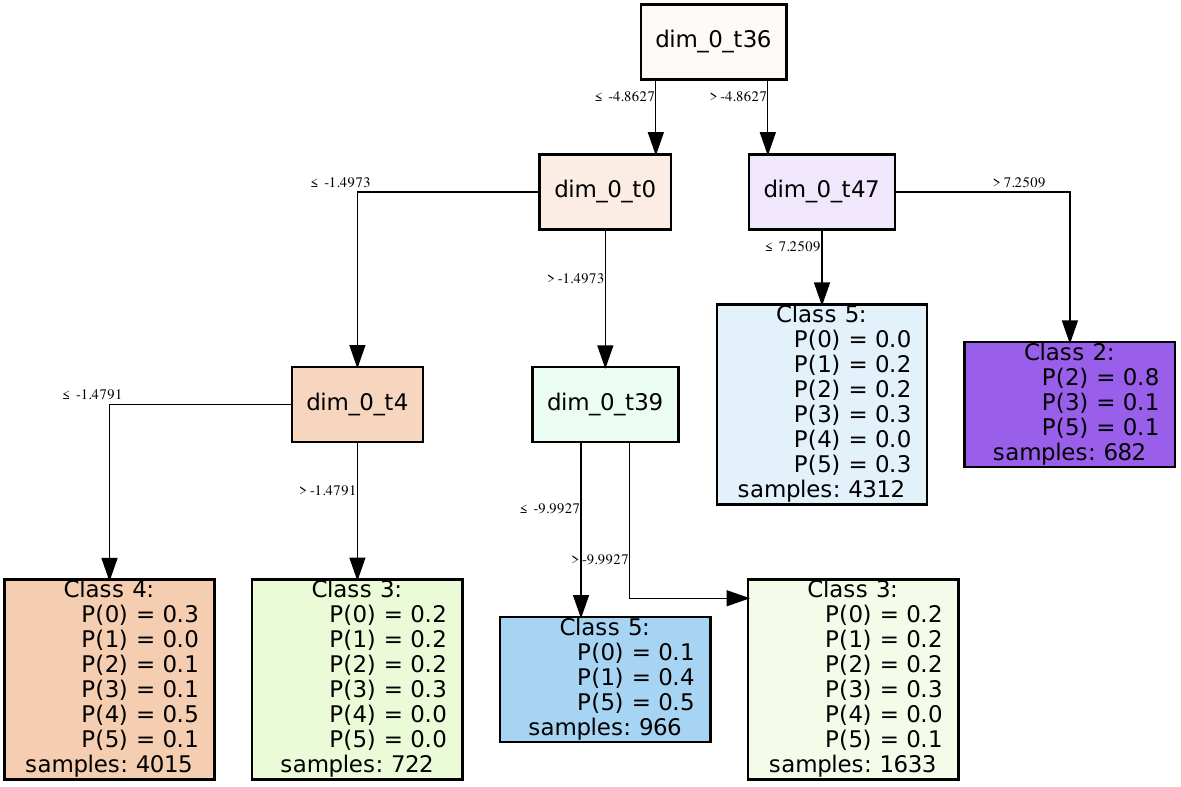}}
    \subfloat[Most relevant parts per time series, one random example per class from the test set \label{fig:instances}]{
        \includegraphics[width=.5\linewidth]{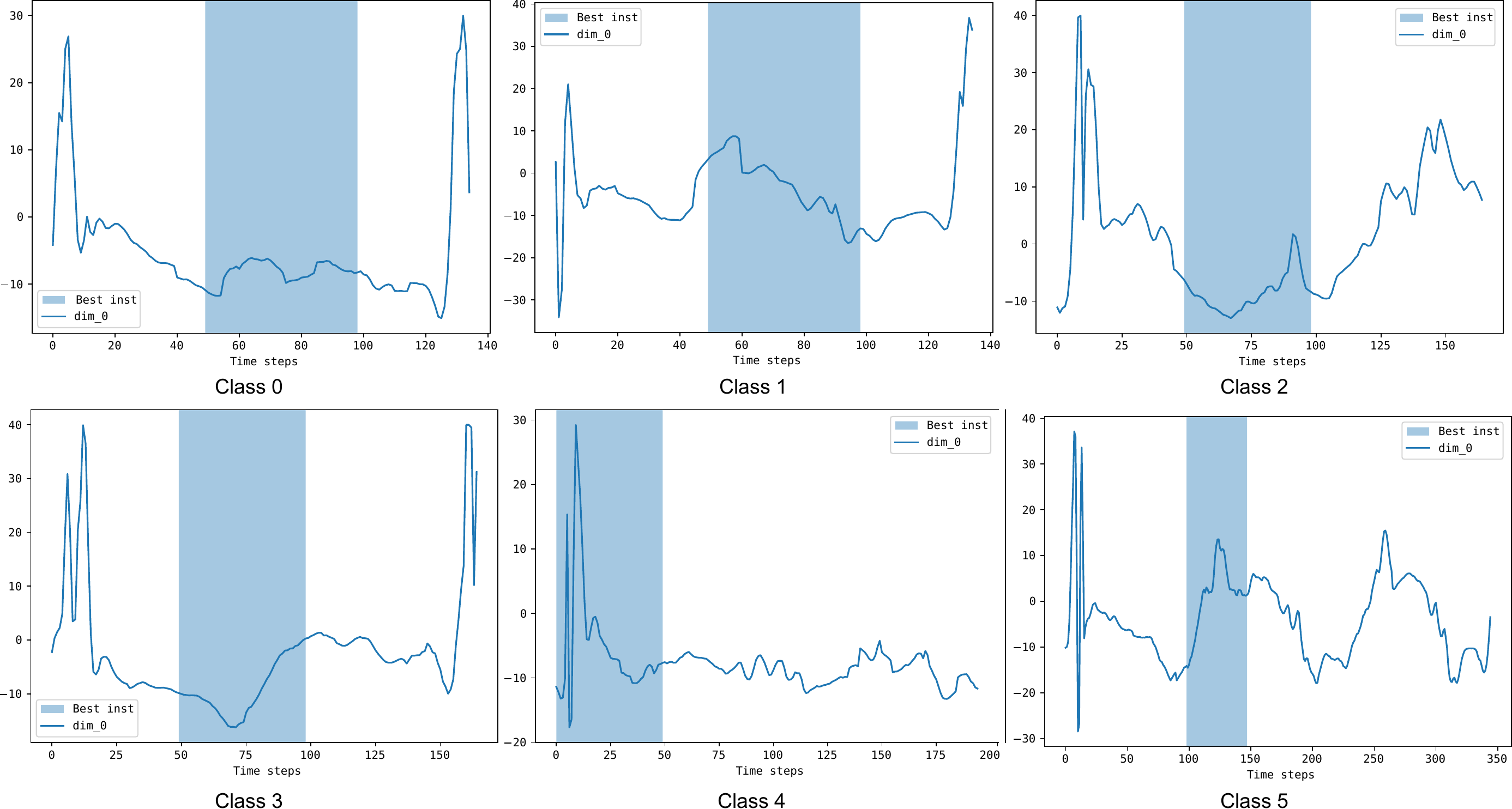}
    }
    \caption{Interpretability in MIHT, taking \textit{GesturePebbleZ1} as example}
\end{figure}

One of the strengths of \ac{MIHT} is its interpretability, which is obtained in two ways. On the one hand, the base classifier generated is composed of a single decision tree, which is naturally interpretable. On the other hand, the designed \ac{MIL} framework works at the instance level to determine which instances maximize the probability of the bag class. 
Figure \ref{fig:tree} illustrates MIHT's interpretability through a representative example from the \textit{GesturePebbleZ1} dataset. The tree structure reveals transparent decision-making, with nodes directly referencing specific variables and time steps without preprocessing. In this case, the model identifies dimension 0 at the 36th instant as the most discriminative feature, with a threshold of $-4.86$
. The MIL framework's ability to analyze sub-sequences results in more compact tree structures compared to whole-series approaches.
The model's dimensionality is further controlled through its leaf node classification system explained in Section \ref{sec:prop:eval}. 
This also explains why in Figure \ref{fig:tree}, in the leaf nodes, there are classes that do not appear in the majority class section. These classes still could be predicted at those nodes, but with lower probability
.


The MIL framework's significance extends to identifying the most relevant series segments for classification. As discussed in Section \ref{sec:prop:bestk}, when classifying an unknown \ac{MIL} bag, the model leverages the leaf nodes' probabilistic outputs to identify the instance that maximizes the predicted class probability. This instance corresponds to the time series segment most characteristic, providing insights into class-specific temporal patterns.
Figure \ref{fig:instances} demonstrates MIHT's pattern recognition capabilities using the \textit{GesturePebbleZ1} dataset, which contains 6 classes with series averaging 219 time steps. With $\omega$ set to 21\%, the model analyzes instances of 46 time steps. The visualization shows one random example per class, highlighting the most representative segments that maximize $\mathcal{L}$. The analysis reveals that most classes exhibit distinctive patterns in the central region of the series
.
It should be noted that the datasets in the UCR/UEA repository do not provide information beyond the general label for the whole series, so the relevance of the marker spans by \ac{MIHT} can not be validated.

\section{Conclusions}\label{sec:conclusions}

This paper has presented \ac{MIHT}, a proposal for multivariate with variable length \ac{TSC} with three main significant innovations. Firstly, it pioneers the use of \acp{IDT} in \ac{TSC}, traditionally applied in data stream learning for their lightning, efficient, and dynamical adaptations to the temporal variations, which has needed the creation of a new learning framework.
Second, it represents the initial effort in combining \ac{MIL} with \ac{IDT}, requiring the adaptation of the paradigm to capture temporal information. \ac{MIL} allows learning the concept that characterizes the series by ignoring non-representative sections. This is carried out by treating the time series as bags with multiple ordered instances. Lastly, this work introduces a new optimization function based on information maximization processes in \acp{IDT}, reinforcing critical segments of the time series.
The experimental results on 28 datasets widely used in \ac{TSC} proved 
that \ac{MIHT} is competitive against 11 state-of-the-art models for \ac{TSC} encompassing problems with variable length. The interpretability properties of the model were also analyzed, which make \ac{MIHT} a valuable approach for identifying both variables and time spans most relevant to the classification.

\begin{credits}
\subsubsection{\ackname} This research was funded by the TED2021-132702B-C22 project of Spanish Ministry of Science and Innovation, by the ProyExcel-0069 project of Andalusian Department of the University, Research and Innovation Andalucian, and by the doctoral grant FPU19/03924 of the Spanish Ministry of Universities.

\end{credits}
%
%
%

\bibliographystyle{splncs04}
\bibliography{refs}

@article{Bagnall2017,
author = {Bagnall, Anthony and Lines, Jason and Bostrom, Aaron and Large, James and Keogh, Eamonn},
journal = {Data Min Knowl Disc},
number = {3},
pages = {606--660},
title = {{The great time series classification bake off: a review and experimental evaluation of recent algorithmic advances}},
volume = {31},
year = {2017}
}

@incollection{Bostrom2017,
author = {Bostrom, Aaro and Bagnall, Anthony},
booktitle = {17th Int Conf on Big Data Analytics and Knowledge Discovery},
pages = {23--46},
title = {{Binary shapelet transform for multiclass time series classification}},
year = {2017},
}

@article{Chen2023,
author = {Chen, Long and Lian, Cheng and Zeng, Zhigang and Xu, Bingrong and Su, Yixin},
journal = {Inform Sciences},
title = {{Cross-modal multiscale multi-instance learning for long-term ECG classification}},
volume = {643},
pages = {1--15},
year = {2023}
}

@article{Dempster2020,
author = {Angus Dempster and François Petitjean and Geoffrey I. Webb},
journal = {Data Min Knowl Disc},
pages = {1454-1495},
title = {ROCKET: exceptionally fast and accurate time series classification using random convolutional kernels},
volume = {34},
year = {2020},
}

@inproceedings{Dennis2018,
author = {Dennis, Don Kurian and Pabbaraju, Chirag and Simhadri, Harsha Vardhan and Jain, Prateek},
booktitle = {32nd Conf on Neural Information Processing Systems},
title = {{Multiple Instance Learning for Efficient Sequential Data Classification on Resource-constrained Devices}},
year = {2018}
}

@inproceedings{Domingos2000,
author = {Domingos, Pedro and Hulten, Geoff},
booktitle = {6th ACM SIGKDD Intern Conf on Knowledge Discovery and Data Mining},
title = {{Mining High-Speed Data Streams}},
pages = {71--80},
year = {2000}
}

@inbook{Faouzi2023,
author = {Johann Faouzi},
booktitle = {Time Series Analysis - Recent Advances, New Perspectives and Applications},
title = {Time Series Classification: A review of Algorithms and Implementations},
year = {2024}
}

@article{Fawaz2020,
author = {Hassan Ismail Fawaz and Benjamin Lucas and Germain Forestier and Charlotte Pelletier and Daniel F. Schmidt and others},
journal = {Data Min Knowl Disc},
title = {InceptionTime: Finding AlexNet for time series classification},
volume = {34},
pages = {1936-1962},
year = {2020},
}

@inproceedings{Guan2016,
author = {Guan, Xinze and Raich, Raviv and Wong, Weng Keen},
booktitle = {33rd Intern Conf on Machine Learning},
title = {{Efficient multi-instance learning for activity recognition from time series data using an auto-regressive hidden markov model}},
year = {2016},
pages = {3452--3473},
}

@article{Han2023,
author = {Han, Haozhan and Lian, Cheng and Zeng, Zhigang and Xu, Bingrong and Zang, Junbin and Xue, Chenyang},
journal = {Knowl-based Syst},
title = {{Multimodal multi-instance learning for long-term ECG classification}},
volume = {270},
pages = {1--12},
year = {2023}
}

@inproceedings{Janakiraman2018,
author = {Janakiraman, Vijay Manikandan},
booktitle = {24th ACM SIGKDD on Knowl Disc and Data Mining},
pages = {406--415},
title = {{Explaining aviation safety incidents using deep temporal multiple instance learning}},
year = {2018}
}

@article{Kate2016,
author = {Kate, Rohit J.},
journal = {Data Min Knowl Disc},
number = {2},
title = {{Using dynamic time warping distances as features for improved time series classification}},
volume = {30},
pages = {283--312},
year = {2016}
}

@article{Loning2019,
  title={sktime: A unified interface for machine learning with time series},
  author={L{\"o}ning, Markus and Bagnall, Anthony and Ganesh, Sajaysurya and Kazakov, Viktor and Lines, Jason and Kir{\'a}ly, Franz J},
  journal={arXiv preprint},
pages = {1--9},
  year={2019}
}

@article{Middlehurst2021,
author = {Middlehurst, Matthew and Large, James and Flynn, Michael and Lines, Jason and Bostrom, Aaron and Bagnall, Anthony},
journal = {Mach Learn},
number = {11-12},
pages = {3211--3243},
title = {{HIVE-COTE 2.0: a new meta ensemble for time series classification}},
volume = {110},
year = {2021}
}

@inproceedings{Ruiz-Munoz2015,
author = {Ruiz-Mu{\~{n}}oz, J. F. and Orozco-Alzate, Mauricio and Castellanos-Dominguez, G.},
booktitle = {24th Intern Joint Conf on Artificial Intelligence},
pages = {2632--2638},
title = {{Multiple instance learning-based birdsong classification using unsupervised recording segmentation}},
year = {2015}
}

@inproceedings{Schafer2017,
author = {Sch{\"{a}}fer, Patrick and Leser, Ulf},
booktitle = {3rd ECML PKDD Worksh on Advanc Analyt and Learn on Temp Data},
pages = {1--11},
title = {{Multivariate Time Series Classification with WEASEL+MUSE}},
year = {2017}
}

@article{Stikic2011,
author = {Stikic, Maja and Larlus, Diane and Ebert, Sandra and Schiele, Bernt},
journal = {IEEE T Pattern Anal},
number = {12},
pages = {2521--2537},
title = {{Weakly supervised recognition of daily life activities with wearable sensors}},
volume = {33},
year = {2011}
}

@article{Waqas2024,
   author = {Muhammad Waqas and Syed Umaid Ahmed and Muhammad Atif Tahir and Jia Wu and Rizwan Qureshi},
   journal = {Knowl-based Syst},
   title = {Exploring Multiple Instance Learning (MIL): A brief survey},
   volume = {250},
pages = {1--20},
   year = {2024}
}

@article{Wang2021,
author = {Wang, Guochao and Wang, Yu and Sun, Xiaojie},
journal = {IEEE T Instrum Meas},
title = {{Multi-Instance Deep Learning Based on Attention Mechanism for Failure Prediction of Unlabeled Hard Disk Drives}},
pages = {1--9},
volume = {70},
year = {2021}
}

@inproceedings{Zhang2020,
author = {Zhang, Xuchao and Gao, Yifeng and Lin, Jessica and Lu, Chang Tien},
booktitle = {34th AAAI Conf on Artificial Intelligence},
pages = {6845--6852},
title = {{TapNet: Multivariate time series classification with attentional prototypical network}},
year = {2020}
}
%
\end{document}